\documentclass{article}
\usepackage{ijcai26}

\usepackage{times}
\usepackage{soul}
\usepackage{url}
\usepackage[hidelinks]{hyperref}
\usepackage[utf8]{inputenc}
\usepackage[small]{caption}
\usepackage{graphicx}
\usepackage{amsmath}
\usepackage{amsthm}
\usepackage{booktabs}
\usepackage{algorithm}
\usepackage{algorithmic}
\usepackage[switch]{lineno}

\usepackage{tikz}
\usepackage{pgfplots}
\pgfplotsset{compat=1.18}
\usepgfplotslibrary{statistics}
\usetikzlibrary{arrows.meta, positioning, calc, decorations.pathreplacing}
\usetikzlibrary{intersections}
\usepgfplotslibrary{fillbetween}
\usepgfplotslibrary{groupplots}

\usepackage{xcolor}
\usepackage{multicol}
\usepackage{multirow}
\usepackage[table]{xcolor}
\usepackage{enumitem}
\usepackage{tikz}
\usetikzlibrary{shapes.geometric}
\usetikzlibrary{matrix, positioning}
\usetikzlibrary{arrows.meta, positioning, calc, decorations.pathreplacing}
\usepackage{pgfplots}
\pgfplotsset{compat=newest}
\usetikzlibrary{backgrounds}
\usepackage{pgfplotstable} 
\usepackage{pgf-pie}
\usetikzlibrary{patterns}
\usepackage{amssymb}

\definecolor{cvprblue}{rgb}{0.21,0.49,0.74}
\definecolor{julie}{RGB}{202,60,102}
\definecolor{AvoidingLeakage}{RGB}{176,176,176}
\definecolor{remy}{RGB}{0,53,63}
\definecolor{lucile}{RGB}{38,70,83}

\definecolor{softblue}{RGB}{173, 216, 230} 
\definecolor{softgreen}{RGB}{144, 238, 144} 
\definecolor{softpink}{RGB}{255, 182, 193} 
\definecolor{softpurple}{RGB}{0, 158, 115} 
\definecolor{softorange}{RGB}{255, 224, 178} 
\definecolor{mediumblue}{RGB}{46, 37, 133} 
\definecolor{mediumred}{RGB}{220, 60, 60} 

\title{SynCB: A Synergy Concept-Based Model with Dynamic Routing Between Concepts and Complementary Neural Branches}

\author{
Tores Julie$^1$
\and
Sun Rémy$^2$\and
Sassatelli Lucile$^{3}$\and
Ancarani Elisa$^3$\and
Wu Hui-Yin$^4$\and
Precioso Frédéric$^1$\\
\affiliations
$^1$Université Côte d’Azur, CNRS, Inria, I3S\\
$^2$Université Côte d’Azur, Inria, CNRS, I3S\\
$^3$Université Côte d’Azur, CNRS, I3S\\
$^4$Université Côte d’Azur, Inria\\
\emails
julie.tores@inria.fr
}

\begin{document}

\maketitle

\begin{abstract}
Concept-based (CB) models provide interpretability and support test-time human intervention, while standard neural networks (NN) offer strong task performance but little transparency. Prior work has explored hybrid formulations that integrate concepts and additional representations to improve accuracy, often at the cost of human interventions. 
We introduce the \emph{Synergy Concept-Based Model (SynCB)} framework, that combines a CB branch with a complementary neural branch, and a trainable routing module that dynamically selects which branch to use for each input. Unlike prior models, which fuse residual and concept-based predictions, SynCB keeps the two branches distinct and coordinates them through the routing module. Moreover, both branches are learned jointly, allowing information sharing between the complementary neural branch and CB branches through their common backbone. To improve responsiveness to interventions, we further introduce a test-time intervention policy and a corresponding loss. 
Across five datasets and CB benchmarks, SynCB consistently achieves higher task accuracy while remaining more responsive to human interventions, surpassing the full neural baseline by up to 3.9 percentage points and exceeding the strongest competitor in intervention performance by up to 6.43 percentage points.
\end{abstract}

\section{Introduction}
Despite impressive performance in domains like image and video analysis, deep learning models are often perceived as “black boxes,” offering little insight into how predictions are made. This lack of interpretability is a major challenge in high-stakes domains such as healthcare or social sciences, where practitioners need understandable outputs to trust, validate, and correct model predictions safely.

Since their inception with the seminal Concept Bottleneck Models (CBMs) \cite{cbm}, concept-based models (CB)\footnote{CB refers to concept-base, while CBM refers to the specific Concept Bottleneck Model.} 
\cite{ProbCBM,CEM,Unsup_CBM} have aimed to make neural network (NN) decisions interpretable through human understandable concepts (e.g. color, shapes, ...). CBM \cite{cbm} forces the network to predict the presence of concepts (in the "concept bottleneck") and makes the final decision solely on the basis of these concept predictions. This provides both interpretability and a practical mechanism for human-in-the-loop correction, making the class of CB models appealing for high-stakes and safety-critical applications.

To improve adoption by the wider deep learning community and improve model performances, further works have endeavored to re-introduce elements from end-to-end NNs back into the concept bottleneck \cite{CEM,CEM_PCO,Unsup_CBM,ECBM}. On the one hand, CBMs face a fundamental limitation: their predictive performance depends critically on the \emph{completeness of the predefined concept set}, meaning whether perfect knowledge of all concept values deterministically determines the task label. Re-introducing traditional NNs in the decision can help alleviate issues with this reliance on a complete 
concept set \cite{Prom_and_Pit,Havasi_Parbhoo,Margeloiu_2021}. On the other hand, this dilutes many of the benefits brought by this new framework of CB models \cite{AL,IntCEM}. Indeed, such methods weaken the inherent interpretability and intervenability of CB models decisions.

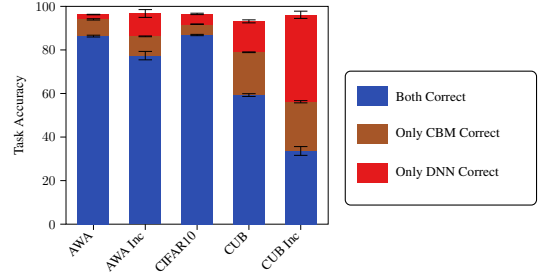
\begin{figure}
    \centering
   
    \resizebox{0.4\textwidth}{!}{
\begin{tikzpicture}

\definecolor{crimson2282628}{RGB}{228,26,28}
\definecolor{darkgray176}{RGB}{176,176,176}
\definecolor{darkslateblue4578176}{RGB}{45,78,176}
\definecolor{lightgray204}{RGB}{204,204,204}
\definecolor{steelblue55126184}{RGB}{55,126,184}
\definecolor{green7717574}{RGB}{77,175,74}
\definecolor{brown1668640}{RGB}{166,86,40}

\begin{axis}[
tick align=outside,
tick pos=left,
x grid style={darkgray176},
xmin=-0.53, xmax=4.53,
xtick style={color=black},
xtick={0,1,2,3,4},
xticklabel style={rotate=45.0,anchor=east},
xticklabels={AWA,AWA Inc,CIFAR10,CUB,CUB Inc},
y grid style={darkgray176},
ylabel={Task Accuracy},
ymin=0, ymax=100,
ytick style={color=black}
]


\draw[draw=none,fill=darkslateblue4578176] (axis cs:-0.3,0) rectangle (axis cs:0.3,86.3540987173716);

\draw[draw=none,fill=darkslateblue4578176] (axis cs:0.7,0) rectangle (axis cs:1.3,77.387809753418);
\draw[draw=none,fill=darkslateblue4578176] (axis cs:1.7,0) rectangle (axis cs:2.3,86.8466675281525);
\draw[draw=none,fill=darkslateblue4578176] (axis cs:2.7,0) rectangle (axis cs:3.3,59.2797160148621);
\draw[draw=none,fill=darkslateblue4578176] (axis cs:3.7,0) rectangle (axis cs:4.3,33.6152344942093);
\draw[draw=none,fill=brown1668640] (axis cs:-0.3,86.3540987173716) rectangle (axis cs:0.3,94.0254539251328);

\draw[draw=none,fill=brown1668640] (axis cs:0.7,77.387809753418) rectangle (axis cs:1.3,86.2357668578625);
\draw[draw=none,fill=brown1668640] (axis cs:1.7,86.8466675281525) rectangle (axis cs:2.3,91.8233342468739);
\draw[draw=none,fill=brown1668640] (axis cs:2.7,59.2797160148621) rectangle (axis cs:3.3,78.9322306712468);
\draw[draw=none,fill=brown1668640] (axis cs:3.7,33.6152344942093) rectangle (axis cs:4.3,56.2133247653643);
\draw[draw=none,fill=crimson2282628] (axis cs:-0.3,94.0254539251328) rectangle (axis cs:0.3,96.244699259599);

\draw[draw=none,fill=crimson2282628] (axis cs:0.7,86.2357668578625) rectangle (axis cs:1.3,96.7314131557941);
\draw[draw=none,fill=crimson2282628] (axis cs:1.7,91.8233342468739) rectangle (axis cs:2.3,96.5433343003194);
\draw[draw=none,fill=crimson2282628] (axis cs:2.7,78.9322306712468) rectangle (axis cs:3.3,93.1193202733994);
\draw[draw=none,fill=crimson2282628] (axis cs:3.7,56.2133247653643) rectangle (axis cs:4.3,96.1339329679807);
\path [draw=black, semithick]
(axis cs:0,85.9433166533045)
--(axis cs:0,86.7648807814387);

\path [draw=black, semithick]
(axis cs:1,75.4554610428905)
--(axis cs:1,79.3201584639454);

\path [draw=black, semithick]
(axis cs:2,86.5479491177942)
--(axis cs:2,87.1453859385108);

\path [draw=black, semithick]
(axis cs:3,58.6469025452579)
--(axis cs:3,59.9125294844663);

\path [draw=black, semithick]
(axis cs:4,31.6011958612761)
--(axis cs:4,35.6292731271424);

\path [draw=black, semithick]
(axis cs:0,93.7291257985023)
--(axis cs:0,94.3217820517632);

\path [draw=black, semithick]
(axis cs:1,86.0557934926937)
--(axis cs:1,86.4157402230313);

\path [draw=black, semithick]
(axis cs:2,91.7207021867432)
--(axis cs:2,91.9259663070045);

\path [draw=black, semithick]
(axis cs:3,78.7663862760352)
--(axis cs:3,79.0980750664584);

\path [draw=black, semithick]
(axis cs:4,55.7000741743067)
--(axis cs:4,56.726575356422);

\path [draw=black, semithick]
(axis cs:0,96.1292045790391)
--(axis cs:0,96.360193940159);

\path [draw=black, semithick]
(axis cs:1,94.9411499009666)
--(axis cs:1,98.5216764106217);

\path [draw=black, semithick]
(axis cs:2,96.2228658679288)
--(axis cs:2,96.8638027327099);

\path [draw=black, semithick]
(axis cs:3,92.4308954316971)
--(axis cs:3,93.8077451151016);

\path [draw=black, semithick]
(axis cs:4,94.4704966065393)
--(axis cs:4,97.7973693294221);

\addplot [semithick, black, mark=-, mark size=5, mark options={solid}, only marks]
table {%
0 85.9433166533045
1 75.4554610428905
2 86.5479491177942
3 58.6469025452579
4 31.6011958612761
};
\addplot [semithick, black, mark=-, mark size=5, mark options={solid}, only marks]
table {%
0 86.7648807814387
1 79.3201584639454
2 87.1453859385108
3 59.9125294844663
4 35.6292731271424
};
\addplot [semithick, black, mark=-, mark size=5, mark options={solid}, only marks]
table {%
0 93.7291257985023
1 86.0557934926937
2 91.7207021867432
3 78.7663862760352
4 55.7000741743067
};
\addplot [semithick, black, mark=-, mark size=5, mark options={solid}, only marks]
table {%
0 94.3217820517632
1 86.4157402230313
2 91.9259663070045
3 79.0980750664584
4 56.726575356422
};
\addplot [semithick, black, mark=-, mark size=5, mark options={solid}, only marks]
table {%
0 96.1292045790391
1 94.9411499009666
2 96.2228658679288
3 92.4308954316971
4 94.4704966065393
};
\addplot [semithick, black, mark=-, mark size=5, mark options={solid}, only marks]
table {%
0 96.360193940159
1 98.5216764106217
2 96.8638027327099
3 93.8077451151016
4 97.7973693294221
};

\end{axis}

\coordinate (legend) at (7.3,4);

\draw[thick, rounded corners] (legend) rectangle ++(5,-3.5);



\def\xoffset{0.3}     
\def\yoffset{-0.9}     
\def\ygap{1}        
\def\rectwidth{0.8} 
\def\rectheight{0.5}

\fill[darkslateblue4578176] ($(legend)+(\xoffset,\yoffset)$) rectangle ++(\rectwidth,\rectheight);
\node[right] at ($(legend)+(\xoffset+\rectwidth+0.1,\yoffset+0.5*\rectheight)$) {Both Correct};

\fill[brown1668640] ($(legend)+(\xoffset,\yoffset-\ygap)$) rectangle ++(\rectwidth,\rectheight);
\node[right] at ($(legend)+(\xoffset+\rectwidth+0.1,\yoffset-\ygap+0.5*\rectheight)$) {Only CBM Correct};

\fill[crimson2282628] ($(legend)+(\xoffset,\yoffset-2*\ygap)$) rectangle ++(\rectwidth,\rectheight);
\node[right] at ($(legend)+(\xoffset+\rectwidth+0.1,\yoffset-2*\ygap+0.5*\rectheight)$) {Only DNN Correct};

\end{tikzpicture}
    }
    \caption{Accuracy gain of CBM models when combined with a ResNet model} 
    \label{fig:motivation}
\end{figure}

We argue that there is in fact promising complementarity between end-to-end networks and CB models: instead of making CB models like end-to-end networks, there are significant benefits to jointly training both types of models. Figure~\ref{fig:motivation} shows that an end-to-end NN (e.g. ResNet \cite{He_2016_CVPR}) and a CB model (e.g. CBM \cite{cbm}, using a ResNet backbone for comparability) have complementary strengths across multiple datasets. Notably, instances where ResNet is correct but the CBM is not highlight opportunities to enhance the CBM by leveraging the strengths of ResNet. This inspires our approach: instead of relying on a single model, integrating complementary predictions from both models can lead to more robust, interpretable and accurate results.

In this work, we propose the \emph{Synergy concept-based model (SynCB)} framework, a concept-model-agnostic framework that reconciles CB models and end-to-end NNs. 
Our approach combines a concept based branch model with a complementary end-to-end NN branch, with both branches sharing a common feature extraction backbone. A routing module dynamically routes samples to either branch, ensuring that predictions remain fully explainable whenever the concept-based branch is selected. 
Moreover, we introduce an intervention policy and an the associated intervention loss to improve the robustness of human-in-the-loop interventions. 

Our contributions can be summarized as follows:

\begin{itemize}
    \item \textbf{SynCB framework}: We propose SynCB, a framework that complements any concept based model with an end-to-end NNs. SynCB achieves this by sharing features between the two models, jointly training a concept based branch and an end-to-end branch. Cooperation between the branches is ensured thanks to our new routing module that attributes each sample to a branch.
    \item \textbf{USI intervention policy}: We introduce USI, a realistic test-time intervention policy that leverages sample-by-sample interventions (or branch attributions) to maximize the benefits of intervention.
    \item \textbf{Extensive evaluation}: We empirically demonstrate that the SynCB framework improves both task accuracy and intervention responsiveness across five datasets and concept benchmarks, outperforming prior state-of-the-art methods that typically excel in only one of these aspects.
\end{itemize}

\section{Related Work}

Beyond refining the use of concepts in the bottleneck \cite{CAV}, later work on CB models have largely followed 3 directions of interest to this work: \\
\noindent$\bullet$ Combine the symbolic reasoning on concepts with traditional end-to-end NN prediction \\
\noindent$\bullet$ Relax the definition of concepts in the bottleneck of concept bottleneck models \\
\noindent$\bullet$ Improve the response of these models to direct intervention on concept detection in the concept bottleneck.

\paragraph{Combing a concept based predictor with an end-to-end neural network predictor}

Most relevant to our work, some further developments have attempted to combine the symbolic CB reasoning of CBMs with a separate traditional end-to-end neural predictor. The Hybrid Post-hoc Concept Bottleneck Model (HP-CBM) \cite{PCBMh} augments CBM frameworks with a residual NN predictor meant to correct shortcomings. This is achieved by training a small NN to yield predictions from the frozen pre-concept bottleneck features of the trained CBM. These predictions are then averaged with the CBM predictions. This design keeps the interpretability of the original CB model while closing some of the performance gap to end-to-end NNs. However, by training the neural residual after the fact on a frozen CB model, HP-CBM does not take full advantage of the complementarity between NNs and CB predictors.

SynCB takes the opposite approach by training a symbolic CB branch and an end-to-end NN branch jointly, and using a learned routing mechanism to force the two branches to complement each other. This leads to better training of the models, acquisition of more general features by the shared feature extractor backbone and the possibility to use the CB branch on its own without losing the benefits of end-to-end representation learning by a NN.

\paragraph{Relaxed concept definition}

Another way to address the limitations of concept bottlenecks is to relax the definition of concepts \cite{CEM,AL}. Hybrid CBMs \cite{Prom_and_Pit,Unsup_CBM} add additional units in the concept bottleneck (serving as unsupervised concepts). Concept Embedding Models (CEMs) \cite{CEM} expand the concept predictions in the bottleneck by instead predicting whole embedding vectors to characterize the presence of each concept. This allows CB models to overcome limitations of strict concept bottlenecks but entangles CB reasoning with black box NN processing. This raises a number of significant issues ranging from worsening the interpretability of decisions to making CB models vulnerable to classical out-of-distribution issues of NNs.

SynCB similarly overcomes the limitations of concept bottlenecks, but through a separate neural path that has no influence on the computation of the clean CB prediction at inference. We therefore retain the ability to control the impact of the end-to-end neural prediction on the process.

\paragraph{Stronger response to interventions on concept detections}

A strong draw of CB models lies in the ability for a human expert to intervene by correcting concept detections at test time, but adding NNs back in the bottleneck typically weakens the impact of interventions. To mitigate this, Intervention aware CEM (IntCEM) \cite{IntCEM} builds on CEMs by explicitly training the model to be receptive to interventions through a learnable intervention policy and an intervention-aware objective that simulates concept corrections during training. Mixture of CEM (MixCEM) \cite{AL} introduces uncertainty-aware embeddings that explicitly regulate leakage, reducing the influence of noisy or ambiguous signals when concepts are unreliable, and thereby protecting interventions from being overridden by leaked information. However, these works specifically focus on combating negative effects of re-introducing NN behavior in the concept bottleneck and do not improve the base performance of models.

To the contrary, our method keeps a pure CB predictor and learns to route samples through a neural predictor to correct possible errors or limitations of the concept bottleneck. This design ensures that interventions remain fully effective in the concept branch, while the complementary branch preserves task accuracy. Furthermore, we re-examine the way interventions are done in CB models and propose new strategies.
\section{SynCB: Reconciling Concept-based Models with Neural Networks}

We propose to synergize a concept-based (CB) symbolic approach with a traditional end-to-end representation learning approach within our new Synergy Concept-Based (SynCB) framework. SynCB consists of an interpretable CB branch complemented by a regular NN branch, allowing the CB decision process to benefit from additional task-relevant information. This synergistic design preserves interpretability and supports concept interventions, while the auxiliary neural branch boosts end-task performance when the CB branch underperforms, as analyzed in Sec. \ref{sec:experiments}.
To jointly preserve task accuracy and intervenability, \textbf{we make the following contribution: we introduce the SynCB framework with three novel elements:}
(i) a trainable routing module at input to route the sample through only one of the two model branches, (ii) a training strategy where we force gradients from both branches to fully benefit the common representation layers, and (iii) a budget-efficient intervention policy where we depart from the random selection of concepts intervened for all test samples as considered in the literature, and instead ensure that a SynCB model can be trained to be highly responsive to well-selected interventions.

In this section, we first provide background on concept-based models (Sec.~\ref{sec:rl_backg}). We then present an overview of the architecture (Sec.~\ref{sec:archi_overview}), followed by a description of the routing module that selects the branch used for each sample at test time (Sec.~\ref{sec:triage}). Next, we introduce our intervention policy (Sec.~\ref{sec:interv_policy}), and finally describe the training procedure and associated losses (Sec.~\ref{sec:archi_training}).

\subsection{Background on CBM and CEM}
\label{sec:rl_backg}
Because our approach integrates a CB branch, we briefly review two canonical formulations that underlie most CB architectures: CBMs \cite{cbm} and CEMs \cite{CEM}.

Both CBMs and CEMs decompose prediction into two stages. First, a backbone network extracts a latent representation that is used by a set of concept predictor branches \( \{cb_i\}_{i=1}^n \), each producing a concept probability \( \hat{p}_i \), where \( n \) denotes the total number of concepts. In CBMs, the task predictor \( f(\cdot) \) operates directly on the concatenated concept probabilities \( [\hat{p}_1, \dots, \hat{p}_n] \) to produce the final prediction \( \hat{y}^{cb} \). In contrast, CEMs introduce a concept embedding \( \hat{c}_i \), constructed from \( \hat{p}_i \), and apply the task predictor to the concatenation of these embeddings \( [\hat{c}_1, \dots, \hat{c}_n] \).
Both formulations support interventions. At both training and testing time, a fraction $p$ of the predicted concept probabilities can be replaced with its corresponding ground-truth $\mathbf{c} = [c_1, \dots, c_n]$, either directly in CBMs or prior to embedding construction in CEMs.

\subsection{Architecture Overview}
\label{sec:archi_overview}

\begin{figure}[ht]
    \hspace*{-1cm}
   
    \scalebox{0.45}{\input{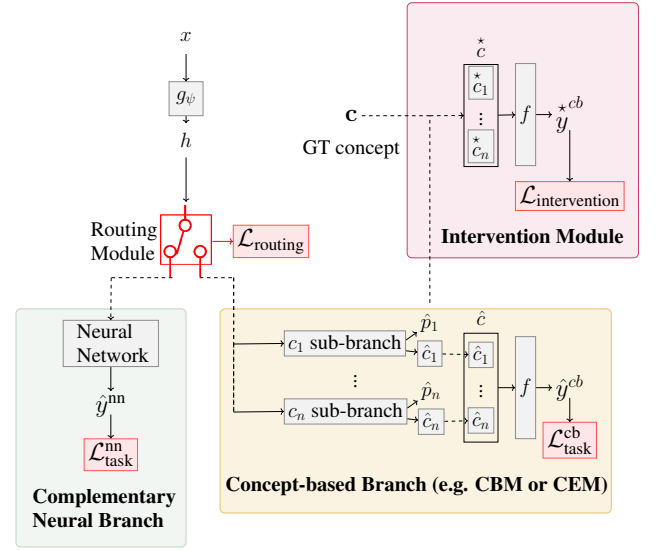}}
    \caption{Overview of SynCB. The shared backbone $g_{\boldsymbol{\psi}}$ maps input $x$ to latent representation $h$, which is fed to both the concept-based and neural branches during training. At test time, the routing module routes each sample to a single branch.}
    \label{fig:architecture}
\end{figure}
 
Figure~\ref{fig:architecture} provides an overview of our framework. An input image $x$ is first mapped by a shared backbone $g_{\boldsymbol{\psi}}$ (typically a ResNet for consistency with prior work \cite{CEM,AL}) into a latent representation: $h = g_{\boldsymbol{\psi}}(x)$,
which is then fed in parallel to two branches that do not share weights: a \textbf{CB branch}, producing interpretable and intervenable predictions, and a \textbf{complementary neural branch}, which focuses on task accuracy. The CB branch can rely on any CB model. In this work we focus on CBMs and CEMs, this branch produces the final prediction $\hat{y}^{cb}$. The neural branch outputs $\hat{y}^{nn}$ through a standard Multi Layer Perceptron. 
At test time, a learnable routing module (Sec.~\ref{sec:triage}), trained jointly with the overall model, selects which branch to use for each sample, and only the chosen branch’s prediction is used as the final output $\hat{y}$. Selecting the CB branch ensures predictions remain fully interpretable and intervenable, whereas selecting the neural branch prioritizes task accuracy. To improve the responsiveness of the CB branch to interventions, we also incorporate an intervention policy and a dedicated loss (Sec.~\ref{sec:interv_policy} and Sec.~\ref{sec:archi_training}).

\subsection{A learnable routing module}
\label{sec:triage}

We introduce this routing module to mitigate the risk, present in models with shortcut pathways, that the task-specific branch dominates the learning process. By using a learnable routing mechanism together with a dedicated loss that encourages correct samples to be routed through the CB branch, we ensure that the complementary branch does not overpower the CB branch.
Our routing module is inspired by the Mixture of Experts paradigm \cite{MOE_1991,Eigen_Ranzato_Sutskever_2014,Fedus_Zoph_Shazeer_2022}, which utilizes a learned gating network to assign each input to a specialized branch.
We define the routing module as a learnable function
\[
R_\mathbf{\phi}: \mathcal{H} \rightarrow [0,1],
\]
parameterized by $\boldsymbol{\phi}$, which is implemented as a small NN with one hidden layer of size 2048. It maps the latent representation $h$ to a routing score $\hat{r} = R_\phi(h)$. The branch selection for sample $i$ is then determined as
\[
b_i = 
\begin{cases}
1, & \text{if }  \hat{r}_i \ge 0.5 \quad \text{(route to concept-based branch)},\\
0, & \text{else }  \text{(route to neural branch)}.
\end{cases}
\]
The module is trained using a standard binary cross-entropy loss, where the target $r_i^* = 1$ if the CB branch predicts the correct task label for sample $i$, and $r_i^* = 0$ otherwise, favoring the CB branch when it is correct, enabling interpretable and intervenable predictions, while defaulting to the complementary neural branch when the CB prediction is incorrect to prioritize task accuracy. We note:  
$$\mathcal{L}_{\text{routing}} = \mathrm{BCE}(\hat{r}, r^*)$$
During training, the routing module is learned but not used to route the samples: all samples are propagated through both branches, and both predictions, $\hat{y}^{cb}$ and $\hat{y}^{nn}$, are used to update the model weights. This ensures that each branch receives sufficient data for effective learning, avoiding underfitting that could result from early routing decisions, and allows the common representation layers $g_{\mathbf{\psi}}$ to receive gradients from both branches.

\subsection{USI: A realistic budget-efficient intervention policy}\label{sec:interv_policy}

The traditional approach to test-time interventions, referred to as Random Concept Intervention (RCI) and adopted in \cite{CEM} and subsequent works \cite{AL}, consists of progressively intervening on a randomly selected $p\%$ of concepts for all test samples. This human-in-the-loop setting requires correcting a fraction of concepts across every sample, which can be unrealistic in practice.

In contrast, we introduce our \textbf{Uncertainty Sample Intervention (USI)} policy, where the human has only to consider $p\%$ of the test samples, but check all the concept of each selected sample. The samples are selected based on their uncertainty. We quantify the uncertainty of a sample by counting how many uncertain concepts it contains. 
For each concept $i$, we define an uncertainty interval parameterized by $\epsilon_i$ and classify the sample as follows:
\[
\begin{cases}
\text{sample is certain for concept } i & \text{if } \hat{p}_i < 0.5 - \epsilon_i, \\
& \text{or } \hat{p}_i > 0.5 + \epsilon_i, \\
\text{sample is uncertain for concept } i & \text{otherwise},
\end{cases}
\]
where $\hat{p}_i$ is the probabilistic indicator of the presence of concept $i$.
The value of $\epsilon_i$ is determined from the distribution of predicted probabilities for concept $i$.  Specifically, we compute the first quartile of the concept probabilities across all samples. 
If the first quartile is high (i.e., superior to 0.2), the probabilities are weakly polarized, indicating high uncertainty for that concept; in this case, we set $\epsilon_i = 0.4$. Otherwise, the concept is considered more certain, and we set $\epsilon_i = 0.2$.
Beyond being more realistic for a human-in-the-loop deployment, our new intervention policy USI shows strong performance gains at constant intervention budget compared to RCI (see Sec. \ref{sec:RQ4}).

\subsection{Training}
\label{sec:archi_training}
We now describe the training procedure of our model, with a focus on the loss functions used to jointly optimize the different components.
The loss components are as follows.

\paragraph{Concept Loss}  
We supervise the CB branch using the same as previous CB papers \cite{cbm,CEM,AL}:\
$$\mathcal{L}_{\text{concepts}} = \mathrm{BCE}(\hat{p}, c)$$

\paragraph{Task Loss}  
Both branches are trained on all samples, with the task loss defined as

\begin{align}
\mathcal{L}_{\text{task}} &= \omega^{cb}\, \mathrm{CE}(\hat{y}^{cb}, y) + \omega^{nn}\, \mathrm{CE}(\hat{y}^{nn}, y) \\
&= \omega^{cb}\, \mathcal{L}_{\text{task}}^{\text{cb}} + \omega^{nn}\, \mathcal{L}_{\text{task}}^{\text{nn}},
\end{align}
subject to $\omega^{cb} + \omega^{nn} = 1$. The weighting coefficients $\omega^{cb}$ and $\omega^{nn}$ allow prioritization of one branch over the other. In practice, we set both weights to 0.5 to balance contributions during training, ensuring that the shared backbone $g_{\mathbf{\psi}}$ receives balanced gradients from both branches. 

\paragraph{Intervenability Loss}  

In order to ensure responsiveness to intervention made at test time, we add an intervention loss term to further encourage the CB branch to leverage ground-truth concept information at test time. This is achieved by feeding $f$ with concept embeddings (or probabilities for CBM) constructed from the ground-truth concept values noted $c^*$.
The resulting intervention loss using embeddings from ground-truth concepts $c^*$ is the following:

\[
\mathcal{L}_{\text{intervention}} = \mathrm{CE}(f(c^*), y)
\]

\paragraph{Global Loss}  
The total training loss combines all components:  $\mathcal{L} = \lambda_t \mathcal{L}_{\text{task}} + \lambda_c \mathcal{L}_{\text{concept}} + \lambda_r \mathcal{L}_{\text{routing}} + \lambda_i \mathcal{L}_{\text{intervention}}$
.

We find that setting the $\lambda's$ away from 1/4 has little impact, but from analyses on validation sets, we set them to $\lambda_t=1/3$, $\lambda_c=1/3$, $\lambda_r=1/9$, $\lambda_i=2/9$.

\section{Experiments}\label{sec:experiments}

\definecolor{cbpurple}{RGB}{126,41,84}
\begin{table*}[!t]
\resizebox{\textwidth}{!}{
    \centering
\begin{tabular}{lccccc}
\toprule
Method & AWA & AWA Inc & CIFAR10 & CUB & CUB-Inc \\
\midrule
DNN & $ \textcolor{cbpurple}{88.57_{\pm 0.31}} \; / \; \textcolor{gray}{NA} $ & $ \textcolor{cbpurple}{87.94_{\pm 0.14}} \; / \;  \textcolor{gray}{NA} $ & $ \textcolor{cbpurple}{91.57_{\pm 0.21}} \; / \;  \textcolor{gray}{NA} $ & $ \textcolor{cbpurple}{73.47_{\pm 0.26}} \; / \;  \textcolor{gray}{NA} $ & $ \textcolor{cbpurple}{73.54_{\pm 0.57}} \; / \;  \textcolor{gray}{NA} $ \\
\midrule
CBM & $ \textcolor{cbpurple}{91.98_{\pm 0.12}} \; / \; \textcolor{mediumblue}{97.38_{\pm 0.07}} $ & $ \textcolor{cbpurple}{71.52_{\pm 16.89}} \; / \; \textcolor{mediumblue}{93.10_{\pm 1.14}} $ & $ \textcolor{cbpurple}{90.46_{\pm 0.30}} \; / \; \textcolor{mediumblue}{79.31_{\pm 0.04}} $ & $ \textcolor{cbpurple}{67.38_{\pm 0.67}} \; / \; \textcolor{mediumblue}{92.05_{\pm 0.05}} $ & $ \textcolor{cbpurple}{38.06_{\pm 1.89}} \; / \; \textcolor{mediumblue}{94.26_{\pm 0.10}} $ \\
CEM & $ \textcolor{cbpurple}{88.14_{\pm 3.54}} \; / \; \textcolor{mediumblue}{97.04_{\pm 1.01}} $ & $ \textcolor{cbpurple}{87.95_{\pm 5.16}} \; / \; \textcolor{mediumblue}{96.52_{\pm 1.59}} $ & $ \textcolor{cbpurple}{90.48_{\pm 0.16}} \; / \; \textcolor{mediumblue}{79.69_{\pm 0.12}} $ & $ \textcolor{cbpurple}{52.11_{\pm 1.26}} \; / \; \textcolor{mediumblue}{87.90_{\pm 0.38}} $ & $ \textcolor{cbpurple}{49.17_{\pm 5.02}} \; / \; \textcolor{mediumblue}{88.45_{\pm 1.02}} $ \\
ProbCBM & $ \textcolor{cbpurple}{86.49_{\pm 1.23}} \; / \; \textcolor{mediumblue}{96.81_{\pm 0.43}} $ & $ \textcolor{cbpurple}{65.05_{\pm 0.93}} \; / \; \textcolor{mediumblue}{95.61_{\pm 0.55}} $ & $ \textcolor{cbpurple}{78.97_{\pm 1.66}} \; / \; \textcolor{mediumblue}{77.57_{\pm 0.74}} $ & $ \textcolor{cbpurple}{66.17_{\pm 2.10}} \; / \; \textcolor{mediumblue}{94.24_{\pm 0.56}} $ & $ \textcolor{cbpurple}{55.58_{\pm 0.60}} \; / \; \textcolor{mediumblue}{93.54_{\pm 0.13}} $ \\
MixCEM & $ \textcolor{cbpurple}{89.52_{\pm 0.15}} \; / \; \textcolor{mediumblue}{97.38_{\pm 0.05}} $ & $ \textcolor{cbpurple}{86.05_{\pm 0.46}} \; / \; \textcolor{mediumblue}{95.88_{\pm 0.22}} $ & $ \textcolor{cbpurple}{76.61_{\pm 0.96}} \; / \; \textcolor{mediumblue}{73.99_{\pm 0.10}} $ & $ \textcolor{cbpurple}{48.65_{\pm 1.55}} \; / \; \textcolor{mediumblue}{89.55_{\pm 0.38}} $ & $ \textcolor{cbpurple}{35.97_{\pm 0.84}} \; / \; \textcolor{mediumblue}{89.89_{\pm 0.10}} $ \\
HP-CBM & $ \textcolor{cbpurple}{91.41_{\pm 0.09}} \; / \; \textcolor{mediumblue}{91.39_{\pm 0.22}} $ & $ \textcolor{cbpurple}{91.47_{\pm 0.23}} \; / \; \textcolor{mediumblue}{89.59_{\pm 0.43}} $ & $ \textcolor{cbpurple}{87.13_{\pm 2.48}} \; / \; \textcolor{mediumblue}{52.03_{\pm 3.70}} $ & $ \textcolor{cbpurple}{69.84_{\pm 5.54}} \; / \; \textcolor{mediumblue}{86.87_{\pm 0.40}} $ & $ \textcolor{cbpurple}{69.37_{\pm 5.40}} \; / \; \textcolor{mediumblue}{85.00_{\pm 0.19}} $ \\
\midrule
SynCEM & $ \textcolor{cbpurple}{92.23_{\pm 0.12}} \; / \; \textcolor{mediumblue}{97.48_{\pm 0.11}} $ & $ \textcolor{cbpurple}{91.86_{\pm 0.26}} \; / \; \textcolor{mediumblue}{97.24_{\pm 0.04}} $ & $ \textcolor{cbpurple}{91.12_{\pm 0.31}} \; / \; \textcolor{mediumblue}{76.28_{\pm 0.03}} $ & $ \textcolor{cbpurple}{58.60_{\pm 0.41}} \; / \; \textcolor{mediumblue}{87.80_{\pm 0.19}} $ & $ \textcolor{cbpurple}{67.31_{\pm 10.26}} \; / \; \textcolor{mediumblue}{92.62_{\pm 3.55}} $ \\
SynCBM & $ \textcolor{cbpurple}{82.63_{\pm 1.32}} \; / \; \textcolor{mediumblue}{93.43_{\pm 0.09}} $ & $ \textcolor{cbpurple}{88.39_{\pm 1.47}} \; / \; \textcolor{mediumblue}{96.70_{\pm 1.28}} $ & $ \textcolor{cbpurple}{91.49_{\pm 0.12}} \; / \; \textcolor{mediumblue}{76.85_{\pm 0.06}} $ & $ \textcolor{cbpurple}{72.48_{\pm 2.46}} \; / \; \textcolor{mediumblue}{95.09_{\pm 0.38}} $ & $ \textcolor{cbpurple}{64.09_{\pm 1.13}} \; / \; \textcolor{mediumblue}{95.03_{\pm 0.28}} $ \\
\bottomrule
\end{tabular}
}
\caption{
    \textcolor{cbpurple}{Task accuracy} and \textcolor{mediumblue}{Concept Accuracy} reported as mean ± stds (\%) across three seeds.}\label{tab:fidelities}
\end{table*}


\definecolor{darkbrown}{RGB}{213,94,0}

\begin{table*}[ht]

\resizebox{\textwidth}{!}{
    \centering

\begin{tabular}{lccccc}
\toprule
Method & AWA & AWA Inc & CIFAR10 & CUB & CUB Inc \\
\midrule

SynCEM & $ \textcolor{darkbrown}{92.23_{\pm 0.12}} \; / \; \textcolor{softpurple}{92.57_{\pm 0.20}} $ & $ \textcolor{darkbrown}{91.86_{\pm 0.26}} \; / \; \textcolor{softpurple}{92.43_{\pm 0.12}} $ & $ \textcolor{darkbrown}{91.12_{\pm 0.31}} \; / \; \textcolor{softpurple}{91.17_{\pm 0.23}} $ & $ \textcolor{darkbrown}{58.58_{\pm 0.44}} \; / \; \textcolor{softpurple}{62.00_{\pm 0.18}} $ & $ \textcolor{darkbrown}{67.10_{\pm 10.63}} \; / \; \textcolor{softpurple}{69.08_{\pm 7.87}} $ \\
SynCBM & $ \textcolor{darkbrown}{76.35_{\pm 0.40}} \; / \; \textcolor{softpurple}{89.98_{\pm 0.07}} $ & $ \textcolor{darkbrown}{59.19_{\pm 16.71}} \; / \; \textcolor{softpurple}{91.29_{\pm 0.70}} $ & $ \textcolor{darkbrown}{91.49_{\pm 0.12}} \; / \; \textcolor{softpurple}{91.48_{\pm 0.12}} $ & $ \textcolor{darkbrown}{72.48_{\pm 2.46}} \; / \; \textcolor{softpurple}{74.44_{\pm 2.09}} $ & $ \textcolor{darkbrown}{56.29_{\pm 0.91}} \; / \; \textcolor{softpurple}{72.66_{\pm 1.17}} $ \\
\bottomrule
\end{tabular}
}
\caption{
    Task Accuracy for the \textcolor{darkbrown}{Concept-Based} Path and the \textcolor{softpurple}{Complementary Neural} Path reported as mean ± stds (\%) across three seeds.
   }
\label{tab:triage}
\end{table*}
We will study two versions of our SynCB model: SynCEM, where the CB branch is based on a CEM, and SynCBM, where the CB branch is based on a CBM.
To ensure a fair comparison across all baselines, we integrated our code into the GitHub\footnote{https://github.com/mateoespinosa/cem} codebase released by the authors of the CEM and MixCEM papers. 
We follow the same evaluation protocol and report results using the mean and standard deviation, table colors are chosen to be colorblind-friendly.
 Each experiment is run using three different random seeds.

\textbf{Research Questions}
We explore the following questions: \\
\noindent$\bullet$\textbf{RQ1:} How much improvement does SynCB achieve in concept and task performance compared to the baselines? \\
\noindent$\bullet$\textbf{RQ2:} How does the SynCB routing module operate?  How does each branch perform on the final task? \\
\noindent$\bullet$\textbf{RQ3:} How does SynCBM respond to concept interventions? \\
\noindent$\bullet$\textbf{RQ4:} What are the gains of our intervention policy USI? \\
\noindent$\bullet$\textbf{RQ5:} How does each component of SynCEM impact performance?

\textbf{Baselines}
We compare SynCB against a representative set of CB models: the vanilla Concept Bottleneck Model (CBM)~\cite{cbm}, Concept Embedding Models (CEM)~\cite{CEM}, Probabilistic CBMs (ProbCBM)~\cite{ProbCBM}, the post-hoc hybrid CBM~\cite{PCBMh}, and MixCEM~\cite{AL}.
All models share the same ResNet-based backbone architecture \cite{He_2016_CVPR} and are trained using identical optimization settings (optimizer, learning rate, batch size, and number of epochs). We also apply standard data augmentation techniques (e.g., random cropping, flipping, color jitter) during training to improve generalization. Unless otherwise specified, these settings are kept consistent across models to ensure a fair comparison. Configurations and hyperparameters used for each model are detailed in Suppl~\ref{sec:supp_modeldesc}.

\textbf{Datasets}
We evaluate all methods on three standard benchmarks. 
\emph{CUB}~\cite{CUB}, a fine-grained bird classification dataset with 200 classes and 112 human-annotated attributes. The concept set is complete in the sense defined in the introduction.
\emph{AWA}~\cite{awa2}, an animal classification dataset with 50 classes and 85 human-annotated attributes, and likewise provides a concept set that is complete.
\emph{CIFAR10}~\cite{cifar10}, a 10-class image dataset for which we extract 143 unsupervised concepts following~\cite{label_free}. In this case, the concept set is incomplete.
To study robustness to concept incompleteness, we construct incomplete variants of CUB and AWA by randomly subsampling their concepts, following~\cite{AL}, resulting in \emph{CUB Inc} and \emph{AWA Inc}.

\subsection{RQ1 :  How much improvement does SynCB achieve in concept and task performance compared to the baselines?
}

We begin by evaluating the task and concept fidelity of SynCEM and SynCBM using task accuracy and concept accuracy. Our results are shown in Table~\ref{tab:fidelities}.

\textbf{SynCEM and SynCBM demonstrate strong task performance across all datasets.} SynCEM consistently ranks among the best-performing models on every dataset, confirming its overall effectiveness. Importantly, it maintains high accuracy even in incomplete data settings, surpassing the DNN baseline by $3.9$ percentage points (p.p.) on AWA Inc and achieving performance within $1.4$ p.p. of the DNN baseline on CUB Incomplete, which highlights its robustness to missing concepts. On the CUB dataset, the SynCBM variant outperforms SynCEM, which is expected since CBMs are better suited than CEMs when the concept set is complete.

\textbf{Beyond task accuracy, SynCEM and SynCBM also achieve strong concept detection performance.} Both models consistently rank among the best approaches across all five datasets, indicating that their strong task performance is supported by accurate concept predictions rather than reliance on shortcut learning. Although HP-CBM is the closest competing approach in terms of task accuracy, it performs the worst in terms of concept accuracy and, as we show later in RQ2 (Section~\ref{sec:RQ2}), is significantly less responsive to interventions than SynCEM and SynCBM.

\subsection{RQ2: How does the SynCB routing module operate?  How does each branch perform on the final task?}
\label{sec:RQ2}

\begin{figure}
    \centering

    \resizebox{0.48\textwidth}{!}{
\begin{tikzpicture}

\definecolor{darkgray176}{RGB}{176,176,176}
\definecolor{darkorange25512714}{RGB}{255,127,14}

\begin{groupplot}[group style={group size=2 by 1}]
\nextgroupplot[
tick align=outside,
tick pos=left,
title={SynCEM},
x grid style={darkgray176},
xmin=0.5, xmax=5.5,
xtick style={color=black},
xticklabel style={rotate=45.0,anchor=east},
xticklabels={,,AWA, AWA Inc, CIFAR10, CUB, CUB Inc},
y grid style={darkgray176},
ylabel={Probability of routing through the concept path},
ymin=-0.05, ymax=1.05,
ytick style={color=black}
]
\addplot [black]
table {%
0.75 0.999110877513886
1.25 0.999110877513886
1.25 0.999972939491272
0.75 0.999972939491272
0.75 0.999110877513886
};
\addplot [black]
table {%
1 0.999110877513886
1 0.984601383209228
};
\addplot [black]
table {%
1 0.999972939491272
1 0.999999642372131
};
\addplot [black]
table {%
0.875 0.984601383209228
1.125 0.984601383209228
};
\addplot [black]
table {%
0.875 0.999999642372131
1.125 0.999999642372131
};
\addplot [black]
table {%
1.75 0.999236345291138
2.25 0.999236345291138
2.25 0.999951481819153
1.75 0.999951481819153
1.75 0.999236345291138
};
\addplot [black]
table {%
2 0.999236345291138
2 0.987934677600861
};
\addplot [black]
table {%
2 0.999951481819153
2 0.999999165534973
};
\addplot [black]
table {%
1.875 0.987934677600861
2.125 0.987934677600861
};
\addplot [black]
table {%
1.875 0.999999165534973
2.125 0.999999165534973
};
\addplot [black]
table {%
2.75 0.995877951383591
3.25 0.995877951383591
3.25 0.999753534793854
2.75 0.999753534793854
2.75 0.995877951383591
};
\addplot [black]
table {%
3 0.995877951383591
3 0.963216588497162
};
\addplot [black]
table {%
3 0.999753534793854
3 0.999990583658218
};
\addplot [black]
table {%
2.875 0.963216588497162
3.125 0.963216588497162
};
\addplot [black]
table {%
2.875 0.999990583658218
3.125 0.999990583658218
};
\addplot [black]
table {%
3.75 0.760437786579132
4.25 0.760437786579132
4.25 0.902631506323814
3.75 0.902631506323814
3.75 0.760437786579132
};
\addplot [black]
table {%
4 0.760437786579132
4 0.397572167813778
};
\addplot [black]
table {%
4 0.902631506323814
4 0.981852247714996
};
\addplot [black]
table {%
3.875 0.397572167813778
4.125 0.397572167813778
};
\addplot [black]
table {%
3.875 0.981852247714996
4.125 0.981852247714996
};
\addplot [black]
table {%
4.75 0.985722213983536
5.25 0.985722213983536
5.25 0.998482018709183
4.75 0.998482018709183
4.75 0.985722213983536
};
\addplot [black]
table {%
5 0.985722213983536
5 0.310863067805767
};
\addplot [black]
table {%
5 0.998482018709183
5 0.999981009960175
};
\addplot [black]
table {%
4.875 0.310863067805767
5.125 0.310863067805767
};
\addplot [black]
table {%
4.875 0.999981009960175
5.125 0.999981009960175
};
\addplot [darkorange25512714]
table {%
0.75 0.999834537506104
1.25 0.999834537506104
};
\addplot [darkorange25512714]
table {%
1.75 0.999812424182892
2.25 0.999812424182892
};
\addplot [darkorange25512714]
table {%
2.75 0.998987138271332
3.25 0.998987138271332
};
\addplot [darkorange25512714]
table {%
3.75 0.840913981199265
4.25 0.840913981199265
};
\addplot [darkorange25512714]
table {%
4.75 0.994984984397888
5.25 0.994984984397888
};

\nextgroupplot[
scaled y ticks=manual:{}{\pgfmathparse{#1}},
tick align=outside,
tick pos=left,
title={SynCBM},
x grid style={darkgray176},
xmin=0.5, xmax=5.5,
xtick style={color=black},
xticklabel style={rotate=45.0,anchor=east},
xticklabels={,,AWA, AWA Inc, CIFAR10, CUB, CUB Inc},
y grid style={darkgray176},
ymin=-0.05, ymax=1.05,
ytick style={color=black},
yticklabels={}
]
\addplot [black]
table {%
0.75 0.677726328372955
1.25 0.677726328372955
1.25 0.974329054355621
0.75 0.974329054355621
0.75 0.677726328372955
};
\addplot [black]
table {%
1 0.677726328372955
1 0.177203462719917
};
\addplot [black]
table {%
1 0.974329054355621
1 0.999066073894501
};
\addplot [black]
table {%
0.875 0.177203462719917
1.125 0.177203462719917
};
\addplot [black]
table {%
0.875 0.999066073894501
1.125 0.999066073894501
};
\addplot [black]
table {%
1.75 0.527690827846527
2.25 0.527690827846527
2.25 1
1.75 1
1.75 0.527690827846527
};
\addplot [black]
table {%
2 0.527690827846527
2 9.85733000780087e-11
};
\addplot [black]
table {%
2 1
2 1
};
\addplot [black]
table {%
1.875 9.85733000780087e-11
2.125 9.85733000780087e-11
};
\addplot [black]
table {%
1.875 1
2.125 1
};
\addplot [black]
table {%
2.75 0.960674405097961
3.25 0.960674405097961
3.25 0.99558037519455
2.75 0.99558037519455
2.75 0.960674405097961
};
\addplot [black]
table {%
3 0.960674405097961
3 0.820532394051552
};
\addplot [black]
table {%
3 0.99558037519455
3 0.999618700742722
};
\addplot [black]
table {%
2.875 0.820532394051552
3.125 0.820532394051552
};
\addplot [black]
table {%
2.875 0.999618700742722
3.125 0.999618700742722
};
\addplot [black]
table {%
3.75 0.981747642159462
4.25 0.981747642159462
4.25 0.999161541461945
3.75 0.999161541461945
3.75 0.981747642159462
};
\addplot [black]
table {%
4 0.981747642159462
4 0.750816569328308
};
\addplot [black]
table {%
4 0.999161541461945
4 0.999998569488525
};
\addplot [black]
table {%
3.875 0.750816569328308
4.125 0.750816569328308
};
\addplot [black]
table {%
3.875 0.999998569488525
4.125 0.999998569488525
};
\addplot [black]
table {%
4.75 0.839328274130821
5.25 0.839328274130821
5.25 0.997738838195801
4.75 0.997738838195801
4.75 0.839328274130821
};
\addplot [black]
table {%
5 0.839328274130821
5 0.0056185060786083
};
\addplot [black]
table {%
5 0.997738838195801
5 0.99999737739563
};
\addplot [black]
table {%
4.875 0.0056185060786083
5.125 0.0056185060786083
};
\addplot [black]
table {%
4.875 0.99999737739563
5.125 0.99999737739563
};
\addplot [darkorange25512714]
table {%
0.75 0.891645431518555
1.25 0.891645431518555
};
\addplot [darkorange25512714]
table {%
1.75 0.99998676776886
2.25 0.99998676776886
};
\addplot [darkorange25512714]
table {%
2.75 0.986514389514923
3.25 0.986514389514923
};
\addplot [darkorange25512714]
table {%
3.75 0.995494455099106
4.25 0.995494455099106
};
\addplot [darkorange25512714]
table {%
4.75 0.978279113769531
5.25 0.978279113769531
};
\end{groupplot}

\end{tikzpicture}
    }
    \caption{Distribution of samples probabilities to be routed through the concept path.}
    \label{fig:triage_distrib}
\end{figure}
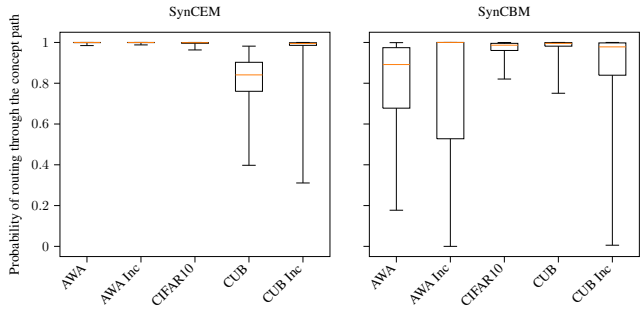

One challenge when using a complementary neural path in CB models is preventing all samples from being routed through the complementary neural path. In Figure~\ref{fig:triage_distrib}, we show the distribution of routing probabilities produced by the routing module and in Table~\ref{tab:triage}, we report the performance of each branch on the final task. The routing module outputs the probability that a sample is routed through the CB branch, probabilities greater than 0.5 (resp. lower) indicate routing through the CB branch (resp. the complementary neural branch).

For SynCEM, the median routing probability is consistently above 0.5, so at least half of the samples go through the CB branch. For AWA, AWA Inc, and CIFAR10, even the minimum routing probability exceeds 0.5, meaning all samples use the CB branch. This aligns with Table~\ref{tab:triage}, where the accuracy gap between branches is small for these datasets (at most 0.57 p.p.) but larger for CUB (3.42 p.p.) and CUB Inc (1.98 p.p.).
In contrast, for SynCBM, although most samples still go through the CB branch, the complementary neural branch outperforms it on AWA, AWA Inc, and CUB Inc by 13.63, 32.1, and 16.37 p.p., respectively, resulting in more samples being routed through it.

\subsection{RQ3: How does SynCBM respond to concept interventions? }
\label{sec:rQ3}
\begin{figure}[h]
    \centering
    \resizebox{0.48\textwidth}{!}{\input{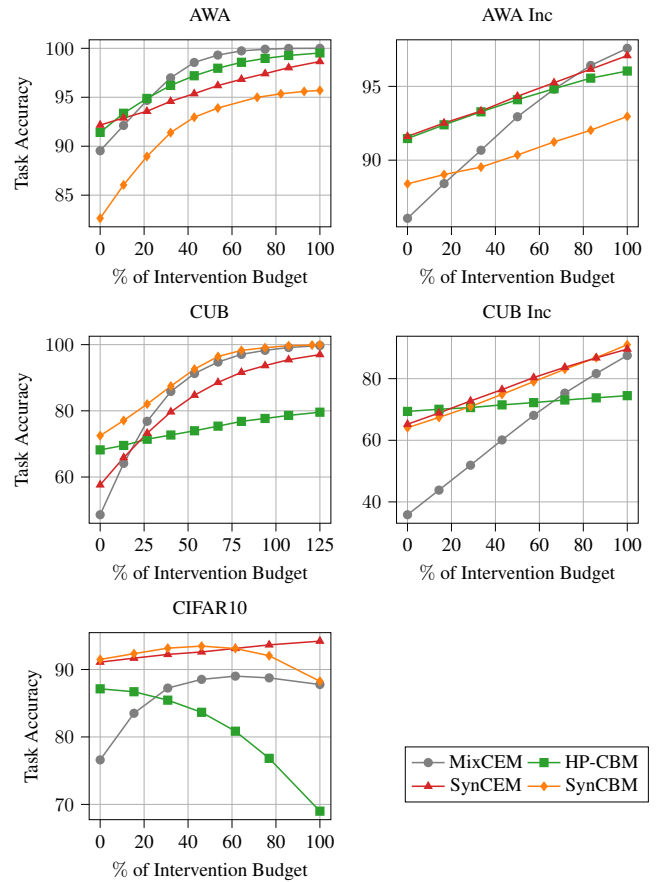}}
    \caption{Task Accuracy as concepts (or group of concepts for task based on CUB and AWA) are intervened following the RCI policy.}
    \label{fig:concept-intervention}
\end{figure}

\begin{table*}[!t]
\centering
\resizebox{\textwidth}{!}{%
\begin{tabular}{l|ccc|ccc|ccc|ccc|ccc}
\toprule
Dataset
& \multicolumn{3}{c|}{AWA}
& \multicolumn{3}{c|}{AWA Inc} 
& \multicolumn{3}{c|}{CIFAR10} 
& \multicolumn{3}{c|}{CUB}
& \multicolumn{3}{c}{CUB Inc} \\
\cmidrule(lr){2-4} \cmidrule(lr){5-7} \cmidrule(lr){8-10} \cmidrule(lr){11-13} \cmidrule(lr){14-16}
Models & {\scriptsize MixCEM} & {\scriptsize SynCEM} & {\scriptsize SynCBM}
& {\scriptsize MixCEM} & {\scriptsize SynCEM} & {\scriptsize SynCBM}
& {\scriptsize MixCEM} & {\scriptsize SynCEM} & {\scriptsize SynCBM}
& {\scriptsize MixCEM} & {\scriptsize SynCEM} & {\scriptsize SynCBM}
& {\scriptsize MixCEM} & {\scriptsize SynCEM} & {\scriptsize SynCBM} \\
\midrule
AUC Diff. & 0.983 & 2.034 & -0.979 & 2.856 & 1.094 & 1.03 & -2.96 & 0.448 & -2.449 &  -13.622 &  -8.163 & -0.462 & 1.529 & 1.224 & 4.988 \\

\bottomrule
\end{tabular}%
}
\caption{
    Difference between the AUC of the USI curve and the RCI curve. }\label{tab:interventioncomp}
\end{table*}

\begin{table*}[ht]
\centering

\resizebox{\textwidth}{!}{
\begin{tabular}{lccccccccc}
\toprule
\multirow{2}{*}{\textbf{Model}} 
& \multicolumn{4}{c}{\textbf{Ablation Elements}} 
& \multicolumn{4}{c}{\textbf{Metrics}} \\
\cmidrule(lr){2-5} \cmidrule(lr){6-9}
 & Interv. Loss & Early Routing & Gradient Concepts & Gradient Residual & Concept Acc. & Task Acc. & CB Acc. & Neural Acc. \\
\midrule
SynCEM         & \checkmark & \texttimes & \checkmark & \checkmark &  $87.80_{\pm 0.19}$ & $58.60_{\pm 0.41}$ & $58.58_{\pm 0.44}$ & $62.00_{\pm 0.18}$ \\
w/o Interv. Loss  & \texttimes & \texttimes & \checkmark & \checkmark & $90.18_{\pm 3.72}$ & $62.97_{\pm 8.45}$ & $62.62_{\pm 8.77}$ & $65.06_{\pm 6.78}$ \\
with Early Routing  & \checkmark & \checkmark & \checkmark & \checkmark &  $87.81_{\pm 0.11}$ & $58.00_{\pm 0.55}$ & $57.94_{\pm 0.63}$ & $61.88_{\pm 0.49}$ \\
w/o Gradient Concepts  & \checkmark & \checkmark & \texttimes & \checkmark &  $86.43_{\pm 0.18}$ & $57.60_{\pm 0.45}$ & $57.46_{\pm 0.43}$ & $60.78_{\pm 0.57}$ \\
w/o  Gradient Residual & \checkmark & \checkmark & \checkmark & \texttimes &  $87.62_{\pm 0.12}$ & $55.52_{\pm 1.59}$ & $55.53_{\pm 1.54}$ & $58.58_{\pm 1.18}$ \\
\bottomrule
\end{tabular}
}
\caption{Ablation Study on the CUB dataset for the SynCEM framework. }
\label{tab:ablation2}
\end{table*}

We investigate the efficiency of concept interventions in our SynCEM and SynCBM models following the Random Concept Intervention (RCI) policy (see Sec.~\ref{sec:interv_policy}), which applies interventions to a randomly selected subset of concepts (or concept groups for the task-based CUB and AWA datasets) across all samples in the batch. Note that intervening on 25$\%$ of the groups is not equivalent to intervening on 25$\%$ of concepts, as groups of concepts may contain varying numbers of concepts. To provide a consistent scale, we define the \emph{intervention budget} such that one unit corresponds to intervening on a single concept for a single sample. Figure~\ref{fig:concept-intervention} shows the RCI curves for all datasets, with task accuracy on the y-axis and the intervention budget on the x-axis. We compare SynCEM and SynCBM to MixCEM, the strongest state-of-the-art competitor for interventions, and HP-CBM, the strongest competitor in terms of task accuracy. Additional graphs and numerical results for other methods are provided in Suppl.~\ref{sec:supp_conceptinteventions}.

SynCEM achieves top performance on CIFAR10, CUB and CUB-Inc. More generally, across all datasets, SynCEM consistently ranks at least second-best, behind MixCEM, which has been specifically designed to perform well under interventions. Notably, the largest observed performance gap between MixCEM and SynCEM is only $1.35$ p.p.
Importantly, the previously strong competitor, HP-CBM, which was competitive in Task Accuracy, struggles to remain effective under interventions, particularly on CIFAR10, CUB, and CUB Incomplete. \textbf{Our model is the only one to achieve both high task accuracy and strong intervention responsiveness simultaneously.}

\subsection{RQ4: How the intervenability varies with the intervention strategy?}\label{sec:RQ4}

\begin{figure}
    \centering
    
    \resizebox{0.48\textwidth}{!}{
\begin{tikzpicture}

\definecolor{darkgray176}{RGB}{176,176,176}
\definecolor{darkorange25512714}{RGB}{255,127,14}
\definecolor{lightgray204}{RGB}{204,204,204}
\definecolor{steelblue31119180}{RGB}{31,119,180}

\begin{groupplot}[group style={group size=2 by 2,
        vertical sep=1.5cm}]
\nextgroupplot[
scaled x ticks=manual:{}{\pgfmathparse{#1}},
height=5cm,
width=6cm,
tick align=outside,
tick pos=left,
title={MixCEM | AWA},
x grid style={darkgray176},
xmajorgrids,
xmin=-5, xmax=105,
xtick style={color=black},
xticklabels={},
y grid style={darkgray176},
ylabel={Task Accuracy},
ymajorgrids,
ymin=85, ymax=101.171166666667,
ytick style={color=black}
]
\addplot [semithick, steelblue31119180, mark=*, mark size=2, mark options={solid}]
table {%
0 89.5333779861576
14.1781647689216 92.9359231971422
28.4620779323116 96.1911140879661
42.7937433513258 98.5532484929672
57.1413261888814 99.5132842152266
71.4655844923368 99.883902656843
85.7865332335211 99.9866041527127
100 100
};
\addplot [semithick, darkorange25512714, mark=square*, mark size=2, mark options={solid}]
table {%
0 89.5155168564412
10 94.3290913150257
20 97.2225943290913
25 98.2451440053583
30 98.9551239115874
40 99.5088189327975
50 99.7365483366823
60 99.7990622906899
70 99.8660415271266
75 99.8883679392722
80 99.8972985041304
90 99.9464166108506
100 100
};

\nextgroupplot[
scaled x ticks=manual:{}{\pgfmathparse{#1}},
scaled y ticks=manual:{}{\pgfmathparse{#1}},
height=5cm,
width=6cm,
tick align=outside,
tick pos=left,
title={SynCEM | AWA},
x grid style={darkgray176},
xmajorgrids,
xmin=-5, xmax=105,
xtick style={color=black},
xticklabels={},
y grid style={darkgray176},
ymajorgrids,
ymin=85, ymax=101.171166666667,
ytick style={color=black},
yticklabels={}
]
\addplot [semithick, steelblue31119180, mark=*, mark size=2, mark options={solid}]
table {%
0 92.1589640544764
14.1781647689216 93.1056039294485
28.4620779323116 94.2710426434472
42.7937433513258 95.36503683858
57.1413261888814 96.3965170797053
71.4655844923368 97.2270596115205
85.7865332335211 98.0218798839027
100 98.6514847064077
};
\addplot [semithick, darkorange25512714, mark=square*, mark size=2, mark options={solid}]
table {%
0 92.1589640544764
10 94.9274391605269
20 96.7269479794597
25 97.3788792141103
30 97.8477338691672
40 98.3612413485153
50 98.544317928109
60 98.5711096226836
70 98.584505469971
75 98.5934360348292
80 98.6202277294039
90 98.6425541415495
100 98.6514847064077
};

\nextgroupplot[
height=5cm,
width=6cm,
tick align=outside,
tick pos=left,
title={MixCEM | CIFAR10},
x grid style={darkgray176},
xlabel={$\%$ of Intervention Budget},
xmajorgrids,
xmin=-5, xmax=105,
xtick style={color=black},
y grid style={darkgray176},
ylabel={Task Accuracy},
ymajorgrids,
ymin=75.4055, ymax=101.171166666667,
ytick style={color=black}
]
\addplot [semithick, steelblue31119180, mark=*, mark size=2, mark options={solid}]
table {%
0 76.6
7.69230769230769 80.37
15.3846153846154 83.51
23.0769230769231 85.9666666666667
30.7692307692308 87.2433333333333
38.4615384615385 88.0966666666667
46.1538461538462 88.5333333333333
53.8461538461538 89.0066666666667
61.5384615384615 89.0133333333333
69.2307692307692 88.7866666666667
76.9230769230769 88.7533333333333
84.6153846153846 88.5933333333333
92.3076923076923 88.5166666666667
100 87.77
};
\addplot [semithick, darkorange25512714, mark=square*, mark size=2, mark options={solid}]
table {%
0 76.5766666666667
10 79.2433333333333
20 81.11
25 81.79
30 82.4466666666667
40 83.8666666666667
50 84.9966666666667
60 86.27
70 87.05
75 87.3166666666667
80 87.4966666666667
90 87.7533333333333
100 87.7833333333333
};

\nextgroupplot[
scaled y ticks=manual:{}{\pgfmathparse{#1}},
height=5cm,
width=6cm,
tick align=outside,
tick pos=left,
title={SynCEM | CIFAR10},
x grid style={darkgray176},
xlabel={$\%$ of Intervention Budget},
xmajorgrids,
xmin=-5, xmax=105,
xtick style={color=black},
y grid style={darkgray176},
ymajorgrids,
ymin=75.4055, ymax=101.171166666667,
ytick style={color=black},
yticklabels={},
legend to name=sl2,
legend columns=2,
legend style={
    at={(5,-6.2)},
    anchor=north,
}
]
\addplot [semithick, steelblue31119180, mark=*, mark size=2, mark options={solid}]
table {%
0 91.09
7.69230769230769 91.38
15.3846153846154 91.68
23.0769230769231 91.9666666666667
30.7692307692308 92.2333333333333
38.4615384615385 92.4733333333333
46.1538461538462 92.6
53.8461538461538 92.8366666666667
61.5384615384615 93.1233333333333
69.2307692307692 93.33
76.9230769230769 93.6466666666667
84.6153846153846 93.8366666666667
92.3076923076923 93.98
100 94.2
};
\addlegendentry{Random Concept Intervention (RCI)}
\addplot [semithick, darkorange25512714, mark=square*, mark size=2, mark options={solid}]
table {%
0 91.09
10 92.0233333333333
20 92.3633333333333
25 92.48
30 92.6333333333333
40 92.93
50 93.21
60 93.5533333333333
70 93.9633333333333
75 94.05
80 94.1266666666667
90 94.1966666666667
100 94.2
};
\addlegendentry{Uncertainty Sample Intervention (USI)}
\end{groupplot}
\ref{sl2}
\end{tikzpicture}}
    \caption{Task Accuracy as we intervene following RCI and USI. }
    \label{fig:sample-intervention}
\end{figure}
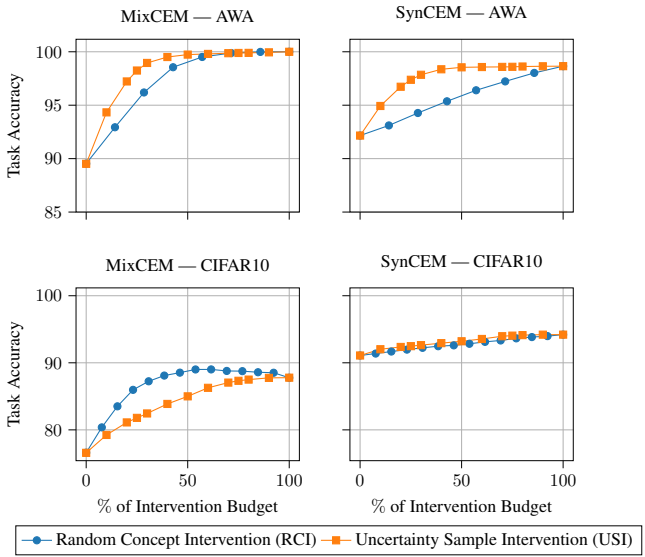

In this section, we investigate the effectiveness of the RCI and USI policies. As introduced in Sec~\ref{sec:interv_policy}, in USI, the human intervenes on all concepts of a selected subset of $p\%$ of test samples, with samples chosen based on their uncertainty. We compare RCI and USI on our two models (SynCEM and SynCBM) and on MixCEM, the best-performing model under RCI. 
To ensure a fair comparison between RCI and USI, we again use the \emph{intervention budget}. Table~\ref{tab:interventioncomp} reports the differences in Area Under the Intervention Evolution Curve for RCI and USI strategies, while Figure~\ref{fig:sample-intervention} provides an illustrative comparison between RCI and USI, with task accuracy on the y-axis and the intervention budget on the x-axis.. 

In Figure~\ref{fig:sample-intervention} we observe that, using the USI strategy allows to reach quicker the intervention plateau. 
RCI performs better on the CUB dataset for MixCEM and on CIFAR-10 for both MixCEM and SynCBM. This can be explained by the fact that these two datasets contain the largest number of concepts (112 for CUB and 143 for CIFAR10), making Uncertainty Sample Intervention (USI) very costly in terms of annotation budget, even though some concepts might already be correct. From the perspective of an expert annotator, who would check all concepts anyway, this cost must be considered when selecting the appropriate intervention strategy. Interestingly, for SynCEM, which is trained by intervening on all concepts of selected samples, the performance drop occurs only for the CUB dataset.
\textbf{In datasets with a human-accessible number of concepts our USI intervention policy brings significant gains compared to RSI.}

\subsection{RQ5 :  How does each component of SynCEM impact performance?}
\label{sec:RQ5}

We conduct a series of ablation experiments to assess the contribution of the main components of our architecture. The results for the CUB dataset are reported in Table~\ref{tab:ablation2}, with additional details provided in Suppl.~\ref{sec:supp_ablation}. Specifically, we analyze: \\
\noindent$\bullet$ The intervenability loss: evaluating performance with and without this loss part \\
\noindent$\bullet$ Gradient flow (GF) from the CB branch to the backbone \\
\noindent$\bullet$ GF from the complementary neural branch to the backbone \\
\noindent$\bullet$ Integrating the routing mechanism during training to enable early sample routing.

First, removing the intervenability loss, results in slightly higher performance in some cases, but considering the standard deviation, the results are less stable. \textbf{This demonstrates that using the intervention loss yields the most consistent and reliable performance.}
Next, we ablate gradient flow from each branch independently to evaluate the impact of joint optimization. Disabling gradient propagation from either the CB branch or the complementary neural branch isolates their influence on the shared backbone. The results show that disabling gradients from the CB branch leads to drops in concept and task accuracy of 1.36 and 1 p.p., respectively. Similarly, disabling gradients from the complementary neural branch causes only a minor drop in concept accuracy (0.18 p.p.) but a larger drop in task accuracy (3.08 p.p.), primarily due to the decrease in accuracy for both branches. \textbf{This results confirms that joint training enables complementary learning and more effective backbone optimization.}
Finally, while early routing encourages the specialization of each branch, we observe a small degradation in final task accuracy (-0.60 p.p.), indicating that premature routing slightly harms the learning quality. \textbf{This supports the design choice of keeping routing decisions strictly for the testing phase.}

\section{Conclusion}

In this paper, we argue that concept based models and end-to-end neural networks can complement each other and introduce the SynCB framework to take advantage of this. 
SynCB combines a concept-based branch with a complementary neural branch thanks to a routing module and joint training procedure. 
In line with our policy of routing samples to different branches in SynCB, we also propose a new sample-by-sample intervention policy USI which better reflects realistic use cases for human interventions
Through extensive evaluation, we show that SynCB improves both task accuracy and responsiveness to interventions compared to prior state-of-the-art methods, 
which typically excelled in only one of these aspects. We further demonstrate that USI improves upon the previous approach in settings with a manageable number of concepts.

\bibliographystyle{named}
\bibliography{ijcai26}

\newpage

\begin{onecolumn}

\section*{Supplementary Material}

\subsection{Training Procedure}

We use a batch size of 128 for AWA, AWA Inc, and CIFAR10, and 64 for CUB and CUB Inc. All models are trained for $E = 150$ epochs, except for CIFAR10, where longer training is beneficial and we use $E = 200$. We optimize using SGD with momentum 0.9. When using a pretrained backbone, the learning rate is set to 0.01 with a weight decay of $4 \times 10^{-6}$; for CIFAR10, we use a learning rate of 0.1 with a weight decay of $1 \times 10^{-6}$.

\subsection{Model description}
\label{sec:supp_modeldesc}
In this section, we describe the models and their hyperparameters. When possible, we followed the choice from the original corresponding work or from MixCEM \cite{AL}.\\

\noindent \textbf{Backbones:} Across all experiments, we employ ResNet-based backbones. For the CUB and AWA datasets (including their incomplete variants), we use ImageNet-pretrained models: a ResNet-34 backbone for CUB and a ResNet-18 backbone for AWA. For CIFAR10, which consists of lower-resolution images, we follow prior work and adopt a ResNet architecture with smaller convolutional kernels, equivalent to ResNet-20. \\
 
\noindent \textbf{DNN:} For all datasets except CIFAR10, we add an additional linear layer with $200 + n$ concept neurons before the final classification layer; for CIFAR10, we instead replace the last classification layer of the ResNet. \\

\noindent \textbf{Vanilla CBM:} We jointly train a CBM with sigmoidal probabilities. We set the concept loss weight $\lambda_c = 1$ for AWA and CUB based task, and $\lambda_c = 10$ for CIFAR10. \\

\noindent \textbf{CEM:} When training CEM, we use a concept embedding size $m=16$ for all datasets and we intervene during training on a concept with probability  $p_\text{int} = 0.25$ (as suggested by the authors~\cite{CEM}). We set the concept loss weight $\lambda_c = 5$ for AWA and AWA-inc, $\lambda_c = 10$ CIFAR10, and $\lambda_c = 1$ CUB and CUB-Inc. \\

\noindent \textbf{MixCEM:} We again adopt the exact configuration and hyperparameter choices reported in the original work \cite{AL}. In particular, we use the prior loss weight $\lambda_p=0.1$ for CIFAR10 and $\lambda_p=1$ for the other, the residual dropout probability $p_{\text{drop}}=0.1$ for CUB Inc and $p_{\text{drop}}=0.5$ for the other, and the number of calibration epochs $E_{\text{cal}} = 30$ for Platt scaling \cite{platt_scaling}. All remaining hyperparameters are fixed to the values selected for CEMs. \\

\noindent \textbf{ProbCBM:} We closely follow the hyperparameter settings used in the original ProbCBM work \cite{ProbCBM}. Specifically, we use the Adam optimizer \cite{kingma2014adam} with an initial learning rate of $0.001$ ($0.01$ for CIFAR10). We fix the number of training and inference samples to $50$, apply concept interventions during training with probability $p_{\text{int}} = 0.5$, warm up the model for $5$ epochs, and scale the KL divergence regularizer with $\lambda_{\text{KL}} = 1 \times 10^{-5}$. We additionally use a weight decay of $1 \times 10^{-6}$, a learning rate ten times smaller for non-pretrained weights, and gradient norm clipping at $2$.
We train ProbCBMs sequentially using early stopping for a maximum of $E$ epochs, allocating at most $E/2$ epochs to training the concept encoder and the remaining epochs to training the task predictor. We set the concept embedding dimension $m = 32$ for CUB, and $m = 16$ for the other datasets, while the class embedding dimension $D_y = 128$ for CUB and AWA, and $D_y = 64$ for the others. Following \cite{ProbCBM}, interventions are performed by replacing the sampled concept embeddings with the learned embedding means of their corresponding to the ground-truth concept labels. \\

\noindent \textbf{HP-CBM:}
We train HP-CBM by strictly adhering to the hyperparameter configuration reported in \cite{AL}. Training begins with a black-box neural network for the downstream task, constructed by appending two linear layers to the shared backbone $g_{\boldsymbol{\psi}}$, separated by a leaky ReLU activation. The intermediate layer contains one neuron per concept, while the final layer matches the number of task labels. 
Once the black box task predictor has been trained, a training set of embeddings is extracted by projecting the entire training set to the space of the second-to-last layer of this model. Then, we learn the Concept Activation Vector (CAVs)~\cite{CAV} for concept $c_i$ using the vector perpendicular to the decision boundary of a linear SVM, with $\ell_2$ penalty $C=1$, trained to predict concept $c_i$ from the activations of the second-to-last layer of the black box DNN. After that, the HP-CBM sparse linear classifier is fine-tuned.
Finally, the residual layer, containing as many hidden neurons as many concepts, is fine-tuned. \\

\noindent \textbf{SynCB:} For all tasks, we use the same hyperparameters, assigning loss weights of $\lambda_t = 1/3$ for the task, $\lambda_c = 1/3$ for the concepts, $\lambda_r = 1/9$ for the routing, and $\lambda_i = 2/9$ for interventions. For SynCBM, the concept-based model is set to a CBM. For the SynCEM, the CB branch is implemented as a CEM that does not mix presence–absence embeddings, but instead selects a single embedding based on the concept probability. The complementary branch is consistently a multi-layer perceptron with one hidden layer of size 2048, which is also used for the routing module. For training interventions, we follow CEM by setting the intervention probability to $p_\text{int} = 0.25$ . However, instead of selecting a portion of concepts, we randomly select a portion of samples, in line with our USI test-time intervention policy.

\subsection{Dataset description}

\begin{table*}[ht]
\centering

\resizebox{0.9\textwidth}{!}{
    \centering
    \begin{tabular}{l|ccc|cccc}
			\toprule
			Dataset & Training Samples & Validation Samples & Testing Samples &  Nb Tasks & Nb Concepts & Nb Concept Groups\\
			\midrule
     AWA & 22,393 & 7,464  & 7,465 & 50 & 85 & 28 \\
			AWA Inc & 22,393 & 7,464  & 7,465 & 50 & 9 & 6 \\ 
                CIFAR10 & 40,000 & 10,000 & 10,000  & 10 & 143 & NA \\
			CUB & 4,796 & 1,198 & 5,794  & 200 & 112 & 28 \\
			CUB Inc & 4,796 & 1,198 & 5,794 & 200 & 22 & 7 \\

			\bottomrule
    \end{tabular}
}
\caption{Dataset Description.}
\label{tab:supp_datasetdesc}
\end{table*}

The \textbf{AWA} task is constructed from the Animals with Attributes 2 (AWA2) dataset~\cite{awa2}. This dataset contains 37{,}322 images annotated with 50 animal species, such as elephants, foxes, and horses. In addition to class labels, each image is annotated with 85 binary attributes describing visual and semantic properties, including \emph{black}, \emph{flippers}, \emph{furry}, and \emph{bipedal}. These attributes are organized into semantic groups such as \emph{color}, \emph{fur pattern}, \emph{size}, and \emph{limb shape}.  
The resulting concept set is \emph{complete} with respect to the classification task: perfect knowledge of the concept annotations uniquely determines the associated animal species.

The \textbf{AWA Incomplete} task is derived from the AWA task by randomly selecting 9 concepts, reducing the number of concept groups to 6. The goal of this construction is to obtain a concept set that is incomplete with respect to the final classification task, such that the retained concepts alone are insufficient to perfectly identify the animal species.

The \textbf{CIFAR10} image classification task is constructed from the original CIFAR10 dataset~\cite{cifar10}, in which each image is annotated with one of 10 object classes (e.g., airplane, bird, frog, ship). Since CIFAR10 does not provide concept annotations, we rely on the approach proposed by \cite{label_free} to generate a concept set using large language models. This method consists of prompting GPT-3 with three queries:
(1) \emph{``List the most important features for recognizing something as a \{class\}''},
(2) \emph{``List the things most commonly seen around a \{class\}''}, and
(3) \emph{``Give superclasses for the word \{class\}''}.

A post-processing step is then applied to remove overly long concept names and concepts that are too closely related to the class label itself. Next, a CLIP model is used to compute similarity scores between image embeddings and concept text embeddings. While several binarization strategies exist, we follow~\cite{AL} and annotate an image with a concept if its similarity score exceeds the 50th percentile across the dataset; images below this threshold are not assigned the concept.

The \textbf{CUB} bird classification task, introduced by \cite{cbm,CEM}, is derived from the Caltech-UCSD Birds-200-2011 dataset~\cite{CUB}. It consists of 11{,}788 images, each annotated with one of 200 bird species, which defines the final classification task.

Following \cite{cbm,CEM,AL}, each image is also associated with a set of binary visual concepts. Initially, the dataset contains 312 binary attributes describing properties such as wing color, breast pattern, or wing shape. However, these concept annotations are known to be noisy, as they were collected from crowdworkers rather than bird experts. Moreover, each image was annotated by a single annotator, making majority voting at the image level impossible.

To mitigate annotation noise, majority voting is instead performed at the class (species) level. For each species, a concept is retained if it is present in more than 50$\%$ of the images belonging to that species. As a result, all images of a given species share the same set of concepts, regardless of whether a particular concept is visible in every individual image. It is worth noting that, due to this processing procedure, the concept set of the CUB dataset is \emph{complete}.

After this aggregation step, concepts that are too sparse (i.e., present in fewer than 10 species) are removed, yielding a final set of 112 concepts.

Additionally, concepts are organized into groups based on shared semantic attributes. For example, the concepts
\emph{has breastcolor::blue},
\emph{has breastcolor::brown},
\emph{has breastcolor::iridescent}, and
\emph{has breastcolor::purple}
all belong to the same concept group corresponding to breast color.

The \textbf{CUB Incomplete} task is derived from the CUB task by randomly selecting 25$\%$ of the original concept groups and retaining only the concepts belonging to those groups, corresponding to 7 groups and 22 concepts. This subsampling of concept groups is performed once, prior to training, and is independent of the training seed.
As for AWA, the objective of this construction is to obtain a concept set that is incomplete.

\clearpage
\subsection{Concept Interventions}
\label{sec:supp_conceptinteventions}

\begin{table*}[h]
    \centering

\begin{tabular}{p{2.5cm}lccccc}
\toprule
Dataset & Model & 0.0$\%$ & 25.0$\%$ & 50.0$\%$ & 75.0$\%$ & 100.0$\%$ \\
\midrule
AWA & CBM & $91.98_{\pm 0.12}$ & $94.59_{\pm 0.06}$ & $97.47_{\pm 0.04}$ & $98.86_{\pm 0.16}$ & $99.04_{\pm 0.26}$ \\
 & CEM & $88.14_{\pm 3.54}$ & $91.77_{\pm 2.47}$ & $95.50_{\pm 1.57}$ & $98.30_{\pm 0.94}$ & $99.42_{\pm 0.39}$ \\
 & MixCEM & $89.53_{\pm 0.13}$ & $95.57_{\pm 0.33}$ & $99.07_{\pm 0.20}$ & $99.91_{\pm 0.04}$ & $100.00_{\pm 0.00}$ \\
 & HP-CBM & $91.41_{\pm 0.09}$ & $95.37_{\pm 0.15}$ & $97.66_{\pm 0.21}$ & $98.97_{\pm 0.01}$ & $99.52_{\pm 0.11}$ \\
 & ProbCBM & $86.49_{\pm 1.23}$ & $91.02_{\pm 0.50}$ & $97.08_{\pm 0.04}$ & $99.83_{\pm 0.04}$ & $100.00_{\pm 0.00}$ \\
& SynCBM & $82.63_{\pm 1.32}$ & $89.83_{\pm 0.49}$ & $93.67_{\pm 0.07}$ & $95.16_{\pm 0.02}$ & $95.70_{\pm 0.29}$ \\
 & SynCEM & $92.16_{\pm 0.11}$ & $93.93_{\pm 0.08}$ & $95.87_{\pm 0.13}$ & $97.41_{\pm 0.05}$ & $98.65_{\pm 0.08}$ \\
\midrule
CUB & CBM & $67.38_{\pm 0.67}$ & $72.49_{\pm 0.62}$ & $77.63_{\pm 0.88}$ & $79.93_{\pm 0.98}$ & $79.48_{\pm 1.58}$ \\
& CEM & $52.11_{\pm 1.26}$ & $61.58_{\pm 2.45}$ & $71.21_{\pm 4.02}$ & $78.50_{\pm 4.43}$ & $83.74_{\pm 4.35}$ \\
 & MixCEM & $48.60_{\pm 1.56}$ & $80.27_{\pm 0.74}$ & $93.88_{\pm 0.38}$ & $98.28_{\pm 0.16}$ & $99.71_{\pm 0.26}$ \\
& HP-CBM & $69.84_{\pm 5.54}$ & $73.98_{\pm 7.51}$ & $77.67_{\pm 8.93}$ & $80.71_{\pm 10.47}$ & $82.97_{\pm 11.26}$ \\
& ProbCBM & $66.17_{\pm 2.10}$ & $79.88_{\pm 1.11}$ & $93.64_{\pm 0.54}$ & $98.71_{\pm 0.12}$ & $100.00_{\pm 0.00}$ \\
 & SynCBM & $72.48_{\pm 2.46}$ & $83.65_{\pm 1.71}$ & $95.40_{\pm 0.17}$ & $99.07_{\pm 0.08}$ & $99.97_{\pm 0.01}$ \\
 & SynCEM & $57.62_{\pm 0.73}$ & $75.46_{\pm 1.22}$ & $87.37_{\pm 1.50}$ & $93.66_{\pm 1.99}$ & $97.00_{\pm 2.28}$ \\
\bottomrule
\end{tabular}

\vspace{0.5cm}

\begin{tabular}{p{2.5cm}lccccc}
\toprule
Dataset & Model & 0.0$\%$ & 29.0$\%$ & 57.0$\%$ & 86.0$\%$ & 100.0$\%$ \\
\midrule
CUB Inc & CBM & $38.06_{\pm 1.89}$ & $39.72_{\pm 2.03}$ & $41.97_{\pm 2.51}$ & $44.06_{\pm 2.68}$ & $45.46_{\pm 2.99}$ \\
 & CEM & $49.17_{\pm 5.02}$ & $56.17_{\pm 4.31}$ & $62.42_{\pm 3.91}$ & $68.02_{\pm 3.10}$ & $70.68_{\pm 2.78}$ \\
 & MixCEM & $35.83_{\pm 0.91}$ & $51.90_{\pm 0.18}$ & $68.08_{\pm 0.16}$ & $81.66_{\pm 0.13}$ & $87.52_{\pm 0.36}$ \\
 & HP-CBM & $69.37_{\pm 5.40}$ & $70.61_{\pm 6.09}$ & $72.23_{\pm 6.97}$ & $73.78_{\pm 7.97}$ & $74.49_{\pm 8.29}$ \\
 & ProbCBM & $55.58_{\pm 0.60}$ & $63.49_{\pm 0.53}$ & $73.66_{\pm 0.40}$ & $83.37_{\pm 0.31}$ & $87.67_{\pm 0.01}$ \\
& SynCBM & $64.09_{\pm 1.13}$ & $70.94_{\pm 0.75}$ & $79.04_{\pm 0.31}$ & $86.82_{\pm 0.02}$ & $91.02_{\pm 0.38}$ \\
& SynCEM & $65.42_{\pm 12.00}$ & $72.91_{\pm 8.70}$ & $80.35_{\pm 6.10}$ & $86.83_{\pm 4.18}$ & $89.60_{\pm 3.47}$ \\
\bottomrule
\end{tabular}

\vspace{0.5cm}

\begin{tabular}{p{2.5cm}lccccc}
\toprule
Dataset & Model & 0.0$\%$ & 33.0$\%$ & 50.0$\%$ & 83.0$\%$ & 100.0$\%$ \\
\midrule
AWA Inc & CBM & $71.52_{\pm 16.89}$ & $68.47_{\pm 15.44}$ & $66.83_{\pm 14.66}$ & $63.79_{\pm 13.55}$ & $62.15_{\pm 13.21}$ \\
& CEM & $87.95_{\pm 5.16}$ & $90.10_{\pm 4.40}$ & $91.39_{\pm 4.08}$ & $93.80_{\pm 3.42}$ & $94.94_{\pm 3.16}$ \\
& MixCEM & $86.06_{\pm 0.53}$ & $90.67_{\pm 0.28}$ & $92.94_{\pm 0.19}$ & $96.42_{\pm 0.19}$ & $97.58_{\pm 0.29}$ \\
& HP-CBM & $91.47_{\pm 0.23}$ & $93.28_{\pm 0.40}$ & $94.10_{\pm 0.41}$ & $95.56_{\pm 0.30}$ & $96.04_{\pm 0.21}$ \\
& ProbCBM & $65.05_{\pm 0.93}$ & $68.09_{\pm 0.49}$ & $69.39_{\pm 0.39}$ & $73.28_{\pm 0.14}$ & $75.36_{\pm 0.46}$ \\
& SynCBM & $88.39_{\pm 1.47}$ & $89.53_{\pm 1.71}$ & $90.35_{\pm 1.92}$ & $92.03_{\pm 2.52}$ & $92.96_{\pm 2.72}$ \\
& SynCEM & $91.61_{\pm 0.28}$ & $93.33_{\pm 0.22}$ & $94.33_{\pm 0.20}$ & $96.16_{\pm 0.10}$ & $97.08_{\pm 0.08}$ \\
\bottomrule
\end{tabular}

\vspace{0.5cm}

\begin{tabular}{p{2.5cm}lccccc}
\toprule
Dataset & Model & 0.0$\%$ & 23.0$\%$ & 46.0$\%$ & 69.0$\%$ & 100.0$\%$ \\
\midrule
CIFAR10 & CBM & $90.46_{\pm 0.30}$ & $91.58_{\pm 0.13}$ & $91.45_{\pm 0.08}$ & $90.24_{\pm 0.04}$ & $86.47_{\pm 0.03}$ \\
 &CEM & $90.48_{\pm 0.16}$ & $91.65_{\pm 0.29}$ & $92.47_{\pm 0.36}$ & $93.13_{\pm 0.47}$ & $93.72_{\pm 0.14}$ \\
& MixCEM & $76.60_{\pm 0.95}$ & $85.97_{\pm 0.63}$ & $88.53_{\pm 0.53}$ & $88.79_{\pm 0.36}$ & $87.77_{\pm 0.36}$ \\
&HP-CBM & $87.13_{\pm 2.48}$ & $86.05_{\pm 4.03}$ & $83.65_{\pm 6.87}$ & $79.03_{\pm 11.72}$ & $69.00_{\pm 20.89}$ \\
& ProbCBM & $78.97_{\pm 1.66}$ & $82.48_{\pm 1.27}$ & $85.10_{\pm 1.10}$ & $87.03_{\pm 0.59}$ & $87.60_{\pm 0.15}$ \\
& SynCBM & $91.49_{\pm 0.12}$ & $92.79_{\pm 0.16}$ & $93.46_{\pm 0.20}$ & $92.71_{\pm 0.07}$ & $88.26_{\pm 0.02}$ \\
& SynCEM & $91.09_{\pm 0.28}$ & $91.97_{\pm 0.18}$ & $92.60_{\pm 0.31}$ & $93.33_{\pm 0.19}$ & $94.20_{\pm 0.10}$ \\
\bottomrule
\end{tabular}
    \caption{Task accuracy is measured as concepts (or groups of concepts, for task-based evaluation on CUB and AWA) are intervened according to the RCI policy.}
    \label{tab:supp_conceptinterv}
\end{table*}
In Table~\ref{tab:supp_conceptinterv}, we report task accuracy as we intervene on concepts or groups of concepts, with one table per dataset (except for a shared table for AWA and CUB). As in the main paper, results are presented in terms of intervention budget, where one unit corresponds to intervening on a single concept for a single sample. Consequently, the thresholds vary across tables. For instance, in the AWA Inc dataset there are only six concept groups, resulting in six intervention steps that do not correspond to standard 25$\%$, 50$\%$ or 75$\%$ threshods. In Fig.\ref{fig:supp_conceptintervtotal}, we display the graphical equivalent. As in the main paper, SynCB shows strong intervention responsiveness, outperforming all other methods.

\begin{figure}[H]
    \centering
    \input{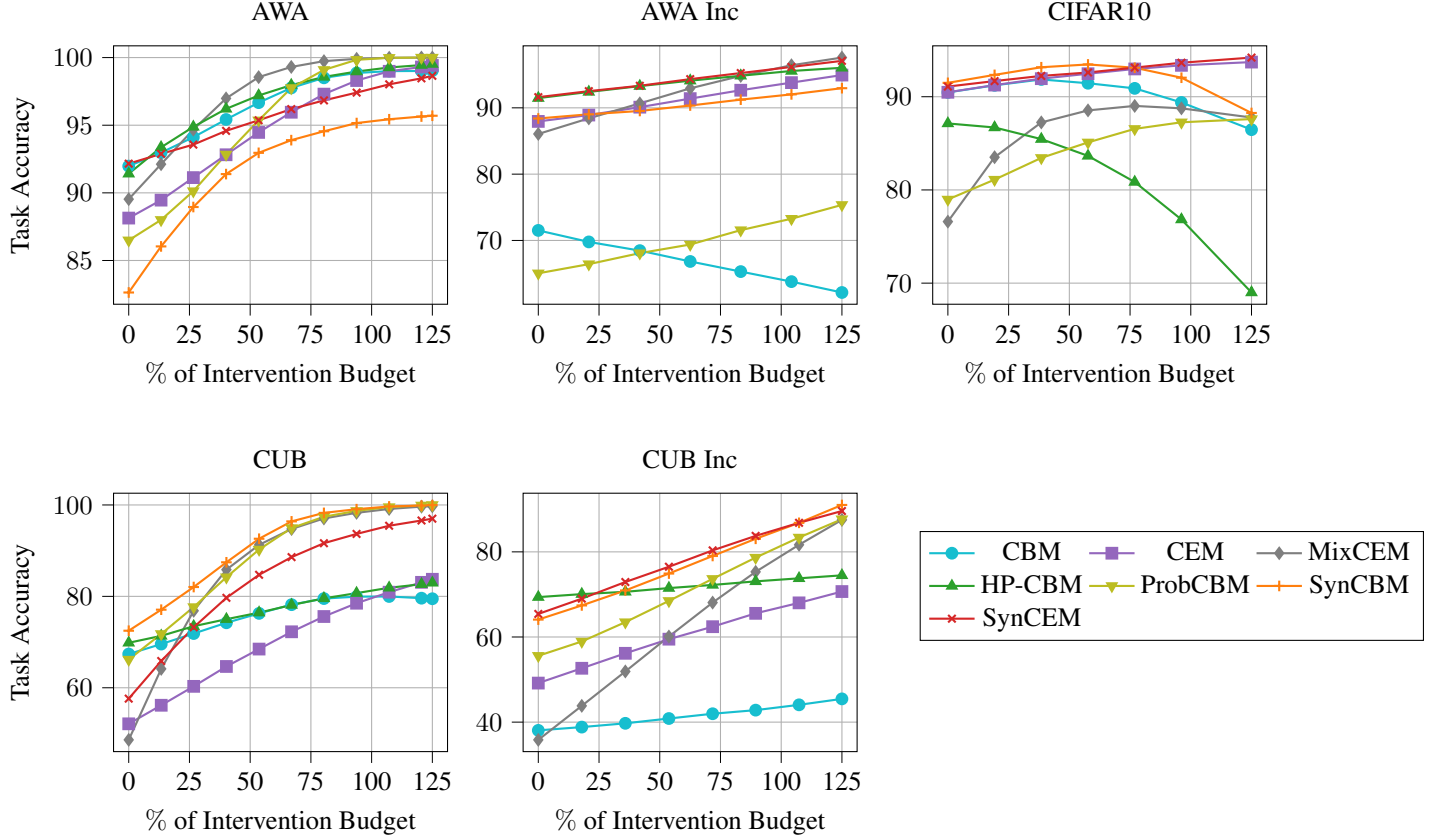}
    \caption{Task Accuracy as concepts (or group of concepts for task based on CUB and AWA) are intervened following the RCI policy.}
    \label{fig:supp_conceptintervtotal}
\end{figure}

\clearpage
\subsection{Ablation Study}
\label{sec:supp_ablation}
We conducted a series of ablation experiments to assess the contribution of the main components of our architecture. The results for all the datasets are reported in Table~\ref{tab:supp_ablationcem}. Specifically, we analyze: \\
\noindent$\bullet$ The intervenability loss: evaluating performance with and without this loss part \\
\noindent$\bullet$ Gradient flow from the CB branch to the backbone \\
\noindent$\bullet$ Gradient flow  from the complementary neural branch to the backbone \\
\noindent$\bullet$ Integrating the routing mechanism during training to enable early sample routing

\begin{table*}

\centering

\begin{tabular}{llcccc}
\toprule
Dataset & Model & Concept Acc. & Task Acc. & CB Acc. & Neural Acc. \\
\midrule
AWA & SynCEM & $97.48_{\pm 0.11}$ & $92.23_{\pm 0.12}$ & $92.23_{\pm 0.12}$ & $92.57_{\pm 0.20}$ \\
& SynCEM w/o Interv. Loss   & $97.63_{\pm 0.12}$ & $92.21_{\pm 0.36}$ & $92.21_{\pm 0.36}$ & $92.41_{\pm 0.19}$ \\
 & SynCEM with Early Routing  & $97.58_{\pm 0.11}$ & $92.05_{\pm 0.27}$ & $92.05_{\pm 0.27}$ & $92.53_{\pm 0.12}$ \\
 & SynCEM w/o Gradient Concepts & $95.99_{\pm 0.11}$ & $92.22_{\pm 0.15}$ & $92.22_{\pm 0.15}$ & $92.44_{\pm 0.06}$ \\
 & SynCEM w/o  Gradient Residual & $97.55_{\pm 0.04}$ & $92.08_{\pm 0.38}$ & $92.08_{\pm 0.38}$ & $92.41_{\pm 0.35}$ \\
\midrule
AWA Inc & SynCEM & $97.24_{\pm 0.04}$ & $91.86_{\pm 0.26}$ & $91.86_{\pm 0.26}$ & $92.43_{\pm 0.12}$ \\
 & SynCEM w/o Interv. Loss   & $97.28_{\pm 0.06}$ & $91.88_{\pm 0.07}$ & $91.88_{\pm 0.07}$ & $92.41_{\pm 0.12}$ \\
 & SynCEM with Early Routing  & $97.22_{\pm 0.01}$ & $91.93_{\pm 0.09}$ & $91.93_{\pm 0.09}$ & $92.48_{\pm 0.18}$ \\
 & SynCEM w/o Gradient Concepts & $96.77_{\pm 0.07}$ & $91.79_{\pm 0.12}$ & $91.79_{\pm 0.12}$ & $92.23_{\pm 0.12}$ \\
 & SynCEM w/o  Gradient Residual & $97.14_{\pm 0.01}$ & $91.75_{\pm 0.06}$ & $91.75_{\pm 0.06}$ & $92.11_{\pm 0.20}$ \\
\midrule
CIFAR10 & SynCEM & $76.28_{\pm 0.03}$ & $91.12_{\pm 0.31}$ & $91.12_{\pm 0.31}$ & $91.17_{\pm 0.23}$ \\
 & SynCEM w/o Interv. Loss   & $76.59_{\pm 0.02}$ & $91.30_{\pm 0.29}$ & $91.30_{\pm 0.29}$ & $91.32_{\pm 0.24}$ \\
 & SynCEM with Early Routing  & $76.21_{\pm 0.04}$ & $91.03_{\pm 0.10}$ & $91.03_{\pm 0.10}$ & $91.07_{\pm 0.14}$ \\
 & SynCEM w/o Gradient Concepts & $75.06_{\pm 0.03}$ & $91.52_{\pm 0.05}$ & $91.52_{\pm 0.05}$ & $91.64_{\pm 0.15}$ \\
 & SynCEM w/o  Gradient Residual & $76.37_{\pm 0.05}$ & $90.34_{\pm 0.11}$ & $90.34_{\pm 0.11}$ & $90.32_{\pm 0.20}$ \\
\midrule
CUB & SynCEM & $87.80_{\pm 0.19}$ & $58.60_{\pm 0.41}$ & $58.58_{\pm 0.44}$ & $62.00_{\pm 0.18}$ \\
 & SynCEM w/o Interv. Loss   & $90.18_{\pm 3.72}$ & $62.97_{\pm 8.45}$ & $62.62_{\pm 8.77}$ & $65.06_{\pm 6.78}$ \\
 & SynCEM with Early Routing  & $87.81_{\pm 0.11}$ & $58.00_{\pm 0.55}$ & $57.94_{\pm 0.63}$ & $61.88_{\pm 0.49}$ \\
 & SynCEM w/o Gradient Concepts & $86.43_{\pm 0.18}$ & $57.60_{\pm 0.45}$ & $57.46_{\pm 0.43}$ & $60.78_{\pm 0.57}$ \\
 & SynCEM w/o  Gradient Residual & $87.62_{\pm 0.12}$ & $55.52_{\pm 1.59}$ & $55.53_{\pm 1.54}$ & $58.58_{\pm 1.18}$ \\
\midrule
CUB-Inc & SynCEM & $92.62_{\pm 3.56}$ & $67.69_{\pm 9.61}$ & $67.36_{\pm 10.18}$ & $69.54_{\pm 7.08}$ \\
 & SynCEM w/o Interv. Loss   & $94.44_{\pm 0.20}$ & $72.11_{\pm 0.99}$ & $72.11_{\pm 0.99}$ & $73.00_{\pm 0.95}$ \\
 & SynCEM with Early Routing  & $94.55_{\pm 0.26}$ & $73.20_{\pm 0.73}$ & $73.20_{\pm 0.73}$ & $74.84_{\pm 0.94}$ \\
 & SynCEM w/o Gradient Concepts & $87.20_{\pm 0.10}$ & $55.64_{\pm 0.43}$ & $54.00_{\pm 0.94}$ & $60.65_{\pm 0.12}$ \\
 & SynCEM w/o  Gradient Residual & $91.38_{\pm 4.31}$ & $63.97_{\pm 12.31}$ & $63.45_{\pm 13.05}$ & $65.89_{\pm 10.48}$ \\
\bottomrule
\end{tabular}
\caption{Ablation Study for the SynCEM model. }
\label{tab:supp_ablationcem}
\end{table*}

Following the observations in Sec.~\ref{sec:RQ5}, we find that the main effects arise from disrupting gradient flows: removing the gradient from the concepts leads to drops in concept accuracy, whereas removing it from the residual branch causes a decrease in task accuracy and reduces the performance of the neural branch.

\end{onecolumn}

\end{document}